\documentclass[10pt,twocolumn,letterpaper]{article}
\usepackage[pagenumbers]{cvpr}
\newcommand{\myparagraph}[1]{\vspace{1mm}\noindent{\bf #1}}

\definecolor{cvprblue}{rgb}{0.21,0.49,0.74}
\usepackage[breaklinks,colorlinks,allcolors=cvprblue]{hyperref}

\def\paperID{18} 
\def\confName{CVPR}
\def\confYear{2026}

\title{How Noisy Poses Break Inverse Dynamics:\\Analysis and Mitigation for Video-Based Joint Torque Estimation}

\author{
Donghyun Kim\textsuperscript{\rm 1, 2}  
\quad 
Chanyoung Kim\textsuperscript{\rm 1, 2}  
\quad 
Eunseo Jeong\textsuperscript{\rm 2}  
\quad 
Youngjoong Kwon\textsuperscript{\rm 1}  
\quad 
Seong Jae Hwang\textsuperscript{\rm 2}  
\\
\textsuperscript{\rm 1}Emory University 
\quad 
\textsuperscript{\rm 2}Yonsei University 
\\
{\tt\small \{donghyun.kim, chanyoung.kim, youngjoong.kwon\}@emory.edu} 
\\
{\tt\small \{eunseo8946, seongjae\}@yonsei.ac.kr}
}

\begin{document}
\maketitle
\begin{abstract}
Recent advances in monocular 3D human pose estimation enable accurate body tracking from video.
However, translating these kinematic estimates into physical quantities, such as joint torques, remains challenging due to \textit{noise amplification} through inverse dynamics.
In this work, we provide a systematic analysis of how pose estimation noise propagates through the inverse dynamics pipeline.
We present three key findings:
(\romannumeral1)~pose noise is amplified by approximately $1{,}000\times$ when computing joint torques via numerical differentiation,
(\romannumeral2)~proximal joints (spine, hips) are up to $10\times$ more sensitive to noise than distal joints (wrists, hands), and
(\romannumeral3)~low-pass filtering before differentiation substantially reduces this amplification.
To enable this analysis, we develop SMPL-Dynamics, a fully differentiable inverse dynamics module for the SMPL body model that requires no external physics simulators.
Our module supports end-to-end gradient computation, and we demonstrate this through differentiable pose refinement, which reduces torque error by 93\% with negligible change in pose.
\end{abstract}

\section{Introduction}
\label{sec:intro}

\begin{figure*}[t]
    \centering
    \includegraphics[width=0.95\textwidth]{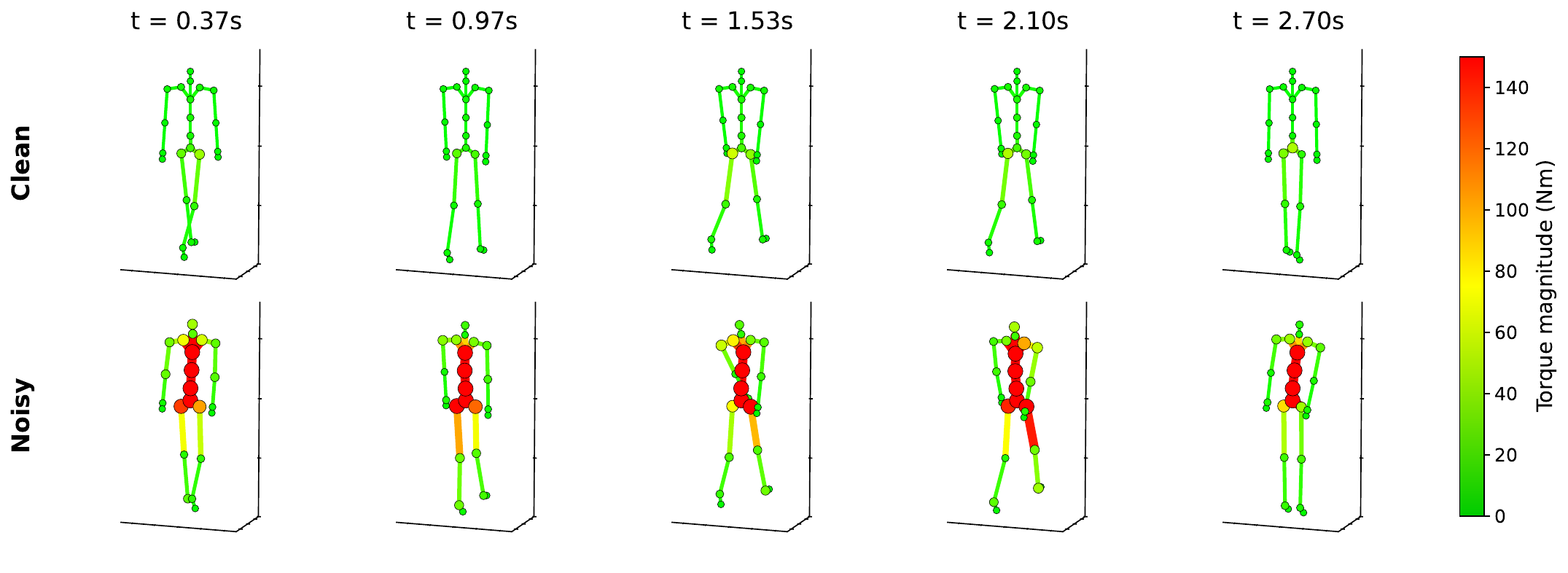}
    \vspace{-10pt}
    \caption{\textbf{Noise amplification in inverse dynamics.} Joint torques visualized as color-coded spheres on the SMPL skeleton during walking (green = low, red = high torque magnitude). \textbf{Top:} Clean motion produces uniformly low, physically plausible torques. \textbf{Bottom:} Adding typical video-based pose noise ($\sigma{=}0.05$\,rad) causes torques to explode at proximal joints (spine, hips), exceeding 150\,Nm, while distal joints remain largely unaffected.}
    \label{fig:3d_walking}
\end{figure*}

Vision-based 3D human pose estimation has progressed rapidly~\cite{shin2024wham, goel2023hmr2, dwivedi2024tokenhmr},
achieving impressive accuracy on standard benchmarks.
However, the computer vision community has largely approached human motion strictly as a problem of \textit{kinematics}, focusing on measuring the spatial positions and joint angles of body parts.
In doing so, it has neglected the \textit{dynamics}, which calculates the underlying physical forces and joint torques that actually generate the observed movements.

Joint torques, ground reaction forces (GRFs), and center-of-mass (CoM) dynamics are essential for biomechanical analysis in sports science~\cite{opencap2023}, clinical rehabilitation, ergonomics, and physics-based character animation~\cite{shimada2020physcap}.
The standard approach to obtain these quantities is \textit{inverse dynamics}: given joint kinematics (\eg, positions, velocities), compute the forces and torques required to produce the observed motion via Newton-Euler equations~\cite{featherstone2008}.

The critical problem is that inverse dynamics requires computing \textit{second-order time derivatives} (\ie, accelerations) of the estimated joint trajectories.
Since video-based pose estimators provide only discrete joint angles per frame, velocities and accelerations must be obtained via numerical differentiation.
A position error $\epsilon$ grows to a velocity error $\epsilon / \Delta t$ after the first derivative and an acceleration error $\epsilon / \Delta t^2$ after the second, where $\Delta t$ is the frame interval.
This quadratic amplification means that even small pose estimation errors, acceptable for kinematic tasks, can produce wildly inaccurate torque estimates (\cref{fig:3d_walking}).
While several works have noted this difficulty~\cite{physpt2024, osdcap2024, uchida2022conclusion, imdy2025}, no work has deeply analyzed \textit{how} and \textit{why} video-based pose error degrades torque estimates (\ie, inverse dynamics).

In this paper, we provide the first systematic analysis of how video-based pose estimation noise propagates through inverse dynamics.
We present a quantitative noise analysis of the video-to-torque pipeline, showing that pose error is amplified by ${\sim}1{,}000\times$ into torque error, that proximal joints (\eg, spine, hips) are up to $10\times$ more sensitive than distal joints (\eg, wrists, hands), and that Butterworth low-pass filtering before differentiation effectively mitigates this effect.
To enable this analysis, we develop \textit{SMPL-Dynamics}, a fully differentiable inverse dynamics module for SMPL~\cite{loper2015smpl} that requires no external physics simulators~\cite{delp2007opensim, todorov2012mujoco}.
We further demonstrate the practical value of differentiability through pose refinement: optimizing poses with a physics-based loss through our module reduces torque error by 93\% with negligible pose change.
\section{Related Work}
\label{sec:related}

\myparagraph{Physics-Based Pose Estimation and Torque Estimation.}
Several works incorporate physics into monocular pose estimation: PhysCap~\cite{shimada2020physcap} adds real-time physics constraints, PhysPT~\cite{physpt2024} refines any kinematic estimate using self-supervised Euler-Lagrange losses, and OSDCap~\cite{osdcap2024} fuses kinematics with dynamics via a neural Kalman filter.
On the torque estimation side, ImDy~\cite{imdy2025} builds a large-scale simulated torque benchmark and trains a data-driven solver for inverse dynamics prediction, OpenCap~\cite{opencap2023} bridges multi-view smartphone videos with OpenSim~\cite{delp2007opensim} for clinical joint moment estimation, and the VID dataset~\cite{vid2025} provides the first benchmark for torque prediction from real monocular images.
However, none of these works systematically analyze \textit{how} and \textit{why} pose noise degrades torque estimates, which is the focus of our work.

\myparagraph{Inverse Dynamics and Body Segment Parameters.}
Classical inverse dynamics uses the Recursive Newton-Euler Algorithm (RNEA)~\cite{featherstone2008} to compute joint forces and torques from kinematics.
Body segment parameters (BSPs), including segment masses, centers of mass, and moments of inertia, are typically sourced from anthropometric tables~\cite{deleva1996}.
Tools like OpenSim~\cite{delp2007opensim} and MuJoCo~\cite{todorov2012mujoco} provide mature inverse dynamics implementations but use skeleton representations that are not directly compatible with SMPL, requiring non-trivial conversion.
Our work bridges this gap by implementing differentiable RNEA directly on the SMPL~\cite{loper2015smpl} kinematic tree.

\myparagraph{Uncertainty in Inverse Dynamics.}
The biomechanics community has long recognized the sensitivity of inverse dynamics to input errors.
Uchida and Seth~\cite{uchida2022conclusion} quantified how \textit{static} errors in marker registration and model scaling propagate through the OpenSim inverse dynamics pipeline, showing that small marker placement uncertainty can produce clinically significant variation in joint moments.
However, their analysis addresses model calibration uncertainty (a constant bias per session) rather than the stochastic, frame-varying noise characteristic of video-based pose estimation.
Our work complements theirs by analyzing a fundamentally different error source: high-frequency pose estimation noise whose spectral characteristics cause severe amplification through numerical differentiation.

\begin{figure*}[t]
    \centering
    \includegraphics[width=\textwidth]{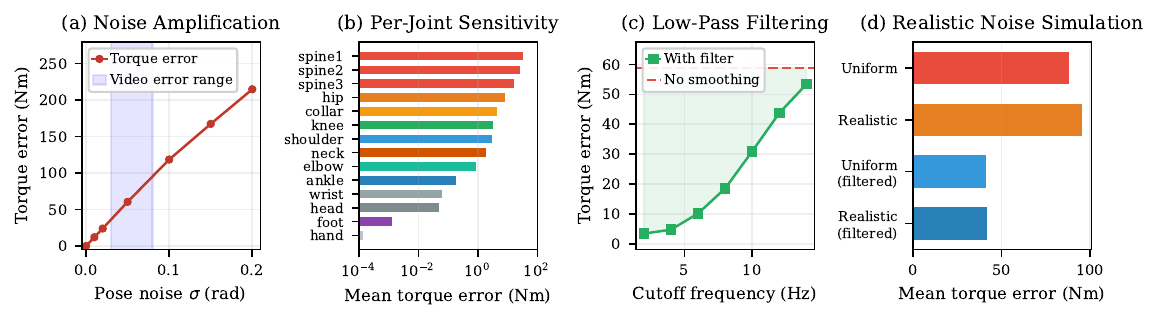}
    \vspace{-20pt}
    \caption{\textbf{Noise amplification analysis.} (a)~Torque error grows approximately linearly with pose noise, with a ${\sim}1{,}000\times$ amplification factor. The blue band indicates the typical video-based estimation error range ($\sigma \approx 0.03$--$0.08$\,rad). (b)~Per-joint sensitivity ranking: proximal joints (spine, hips) are most sensitive due to subtree mass accumulation. (c)~Low-pass filter cutoff trade-off: lower cutoffs remove more noise but distort fast motion. (d)~Torque error under uniform and realistic noise, before and after low-pass filtering.}
    \label{fig:main_results}
\end{figure*}

\section{SMPL-Dynamics Module}
\label{sec:smpl_dynamics}

To analyze noise propagation, we need a differentiable inverse dynamics implementation that operates directly on SMPL~\cite{loper2015smpl} pose parameters.
We build such a module, which takes a sequence of SMPL parameters (root orientation $\boldsymbol{\theta}_0 \in \mathbb{R}^3$, body pose $\boldsymbol{\theta}_{1:23} \in \mathbb{R}^{23 \times 3}$ in axis-angle, and root translation $\mathbf{t} \in \mathbb{R}^3$) and produces joint torques through the following stages.

\myparagraph{(\romannumeral1) Body Segment Parameterization.}
We assign physical properties to each of the 24 SMPL segments using the anthropometric tables of de~Leva~\cite{deleva1996}: segment mass as a fraction of total body mass, center-of-mass location along the segment, and moment of inertia from the radius of gyration.

\myparagraph{(\romannumeral2) Numerical Differentiation.}
Given the 3D position $\mathbf{p}_i(t)$ of joint $i$ at frame $t$, we compute its velocity $\dot{\mathbf{p}}_i$ and acceleration $\ddot{\mathbf{p}}_i$ via central finite differences:
\begin{equation}
    \small
    \dot{\mathbf{p}}_i(t) = \frac{\mathbf{p}_i(t{+}1) - \mathbf{p}_i(t{-}1)}{2\Delta t}
\end{equation}
\begin{equation}
    \small
    \ddot{\mathbf{p}}_i(t) = \frac{\mathbf{p}_i(t{+}1) - 2\mathbf{p}_i(t) + \mathbf{p}_i(t{-}1)}{\Delta t^2}.
\end{equation}
This step is the primary source of noise amplification, since second-order differences scale the input error by $1/\Delta t^2$.
Angular velocities $\boldsymbol{\omega}_i$ and accelerations $\boldsymbol{\alpha}_i$ are similarly obtained from consecutive rotation matrices via the matrix logarithm~\cite{featherstone2008}, and are subject to the same amplification effect.

\myparagraph{(\romannumeral3) Recursive Newton-Euler Algorithm (RNEA).}
We apply RNEA~\cite{featherstone2008} to compute the net force $\mathbf{f}_i$ and torque $\boldsymbol{\tau}_i$ at each joint $i$.
RNEA traverses the kinematic tree from leaves to root, accumulating forces and torques from child joints:
\begin{equation}
    \small
    \mathbf{f}_i = m_i \ddot{\mathbf{p}}_i - m_i \mathbf{g} + \textstyle\sum_{j \in \text{ch}(i)} \mathbf{f}_j 
    \label{eq:newton} 
\end{equation}
\begin{equation}
    \small
    \boldsymbol{\tau}_i = \mathbf{I}_i \boldsymbol{\alpha}_i + \boldsymbol{\omega}_i \times (\mathbf{I}_i \boldsymbol{\omega}_i) + \textstyle\sum_{j \in \text{ch}(i)} (\boldsymbol{\tau}_j + \mathbf{r}_j \times \mathbf{f}_j),
\label{eq:euler}
\end{equation}
where $m_i$ is the segment mass, $\mathbf{g}$ is gravity, $\text{ch}(i)$ denotes the children of joint $i$, $\mathbf{I}_i$ is the segment's inertia tensor, and $\mathbf{r}_j$ is the position vector from joint $i$ to child joint $j$.
The entire pipeline is differentiable, so gradients flow from torque outputs back to pose inputs.

\section{Noise Amplification in Inverse Dynamics}
\label{sec:noise_analysis}

\subsection{Noise Amplification}
\label{subsec:exp_amplification}

We evaluate our module on SMPL walking sequences (30 fps, 120 frames).
We add isotropic Gaussian noise with varying standard deviation $\sigma \in [0, 0.2]$ radian to the SMPL axis-angle pose parameters of a clean walking sequence, then compute torques via our RNEA implementation. 
\cref{fig:main_results}(a) shows the amplification curve.
At $\sigma = 0.05$ radian, a noise level typical of modern video-based pose estimators~\cite{shin2024wham,goel2023hmr2}, the mean torque error is ${\sim}60$\,Nm.
For context, the maximum hip torque during normal walking is ${\sim}30$\,Nm~\cite{deleva1996}, meaning the noise-induced error \textit{exceeds the actual signal}.
The amplification factor is approximately $1{,}000\times$, driven by the $1/\Delta t^2$ scaling of second-order finite differences.
This result explains why prior works on physics-based pose estimation~\cite{physpt2024,osdcap2024,shimada2020physcap} observe that raw kinematics from video are inadequate for downstream physics tasks.

\subsection{Per-Joint Sensitivity}
\label{subsec:exp_sensitivity}

We inject noise into individual joints (one at a time, $\sigma = 0.05$ radian) and measure the resulting total torque error across all joints.
\cref{fig:main_results}(b) reveals a clear hierarchy:

\begin{itemize}
    \item \textbf{Spine joints} are the most sensitive (spine1: 34.4\,Nm, spine2: 26.0\,Nm, spine3: 16.5\,Nm), because rotation errors at the spine propagate to the entire upper body's mass distribution through the kinematic chain.
    \item \textbf{Hip joints} are moderately sensitive (${\sim}$8.6\,Nm), as the lower limbs carry significant mass ($\sim$14\% of body weight per leg~\cite{deleva1996}).
    \item \textbf{Distal joints} contribute minimally ($<$1\,Nm), since they control only small-mass end-effectors.
\end{itemize}

The sensitivity ordering follows the \textit{subtree mass}: joints controlling larger subtrees accumulate more force/torque errors.
This finding has a direct practical implication: \textit{improving trunk estimation accuracy yields the highest return on investment for downstream torque estimation quality.}
Current pose estimators are often evaluated primarily on limb accuracy~\cite{shin2024wham,goel2023hmr2,dwivedi2024tokenhmr}, but our analysis suggests that trunk accuracy should be prioritized for physics-based applications.

\subsection{Low-Pass Filtering}
\label{subsec:low_pass_filtering}

Before computing derivatives, we apply a zero-phase Butterworth low-pass filter to the SMPL pose parameters.
Since second-order differentiation amplifies a frequency component at $f$\,Hz by $f^2$, high-frequency noise is amplified far more severely than low-frequency motion.
Human motion is concentrated below 10\,Hz, while pose estimation noise is broadband, so filtering before differentiation selectively removes the components that would be amplified most.
We evaluate 4th-order zero-phase Butterworth low-pass filters with cutoff frequencies from 2--14\,Hz on noisy pose sequences ($\sigma = 0.05$\,radian).
\cref{fig:main_results}(c) shows the trade-off between noise removal and motion signal preservation.
Very low cutoffs (2--4\,Hz) remove almost all noise but also distort fast movements, since the fundamental frequency of walking is 1--2\,Hz, but harmonics extend to 6--8\,Hz.
We recommend 6\,Hz as a general-purpose default for typical human activities.

\subsection{Realistic Noise Simulation}
\label{subsec:exp_realistic}

Real video-based pose errors are not uniform across joints.
We simulate a WHAM-like~\cite{shin2024wham} noise profile with three characteristics:
\begin{itemize}
    \item \textbf{joint-dependent noise levels}: distal joints (ankles: $\sigma=0.06$, wrists: $\sigma=0.07$) are noisier than proximal joints (pelvis: $\sigma=0.02$, spine: $\sigma=0.03$), matching typical MPJPE distributions.
    \item \textbf{depth-axis amplification}: $2\times$ noise along the camera depth axis, reflecting monocular depth ambiguity.
    \item \textbf{temporal jitter}: frame-to-frame inconsistency that causes flickering.
\end{itemize}
\cref{fig:main_results}(d) shows that realistic noise (95.9\,Nm) produces higher torque error than uniform noise (88.4\,Nm) of equivalent average magnitude.
This is because realistic noise concentrates error on joints that happen to be in the high-sensitivity range (hips, knees), while also introducing depth-correlated errors that affect the root trajectory.
Low-pass filtering (\cref{subsec:low_pass_filtering}) at 6\,Hz reduces both noise models to ${\sim}$42\,Nm, a 55\% reduction.
The remaining error is dominated by low-frequency pose bias that filtering alone cannot remove, motivating the differentiable pose refinement we present in \cref{sec:exp_optimization}                                                  
\begin{table}[t]
\caption{\textbf{Pose refinement results} after 200 iterations of optimization. Torque error drops by 92\% while pose changes negligibly.}
\vspace{-5pt}
\centering
\small
\begin{tabular}{lcc}
\toprule
 & Torque Error (Nm) & Pose Error (rad) \\
\midrule
Noisy & 59.7 & 0.0795 \\
Refined & 4.7 & 0.0787 \\
\midrule
Reduction & 92.2\% & 0.96\% \\
\bottomrule
\end{tabular}
\label{tab:refinement}
\end{table}

\section{Differentiable Pose Refinement}
\label{sec:exp_optimization}

\begin{figure}[t]
    \centering
    \includegraphics[width=\linewidth]{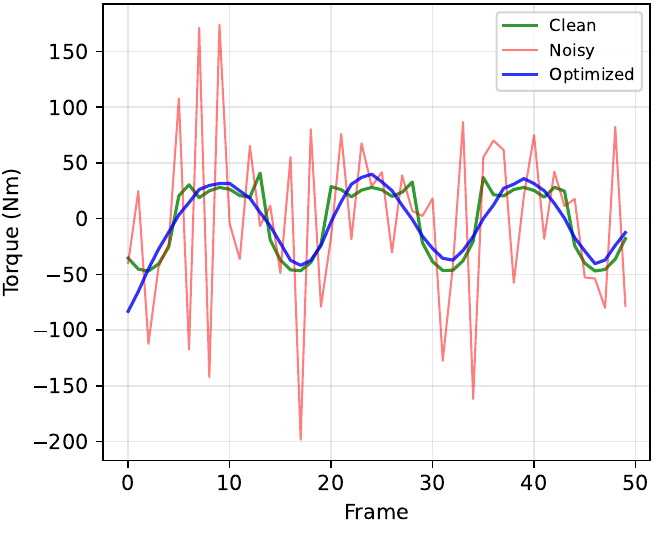}
    \vspace{-20pt}
    \caption{\textbf{Left hip torque before and after refinement.} The optimized trajectory (blue) closely tracks the clean reference (green), while the noisy input (red) oscillates wildly due to noise amplification through inverse dynamics.}
    \label{fig:refinement_torque}
\end{figure}

A key advantage of our module over non-differentiable alternatives such as OpenSim~\cite{delp2007opensim} is that gradients flow from torque outputs back to pose inputs via automatic differentiation.
We demonstrate this by optimizing noisy poses ($\sigma = 0.05$ radian) to minimize a physics-based loss:
\begin{equation}
    \mathcal{L} = \lambda_s \|\ddot{\boldsymbol{\tau}}\| + \lambda_m \text{ReLU}(\|\boldsymbol{\tau}\| - \tau_{\max}) + \lambda_r \|\hat{\boldsymbol{\theta}} - \boldsymbol{\theta}_{\text{noisy}}\|
\end{equation}
where $\hat{\boldsymbol{\theta}}$ denotes the refined pose parameters, $\boldsymbol{\theta}_{\text{noisy}}$ the initial noisy estimate, $\boldsymbol{\tau}$ the joint torques computed by our module from $\hat{\boldsymbol{\theta}}$, and $\ddot{\boldsymbol{\tau}}$ the second time derivative of $\boldsymbol{\tau}$. Each term enforces a different physical prior:
\begin{itemize}
    \item \textbf{Torque smoothness}: $\lambda_s \|\ddot{\boldsymbol{\tau}}\|$ penalizes rapid changes in torque over time, encouraging temporally smooth dynamics ($\lambda_s{=}10$).
    \item \textbf{Physiological magnitude limit}: $\lambda_m \text{ReLU}(\|\boldsymbol{\tau}\| - \tau_{\max})$ penalizes torques exceeding the physiological limit $\tau_{\max}{=}100$\,Nm ($\lambda_m{=}1$).
    \item \textbf{Pose regularizer}: $\lambda_r \|\hat{\boldsymbol{\theta}} - \boldsymbol{\theta}_{\text{noisy}}\|$ keeps the refined pose close to the original estimate, preventing large kinematic deviations ($\lambda_r{=}5$).
\end{itemize}
Using Adam optimization for 200 iterations, torque error drops from 59.7\,Nm to 4.7\,Nm (92\% reduction) while the pose changes negligibly (\Cref{tab:refinement}).
\cref{fig:refinement_torque} shows the effect on the left hip joint, where the optimized torque trajectory closely tracks the clean reference.
This remarkable result shows that the solution space contains physically plausible poses \textit{very close} to the noisy estimates, and that our differentiable inverse dynamics can efficiently find them.

\section{Conclusion}
\label{sec:conclusion}

We present a systematic analysis of how video-based pose estimation noise propagates through inverse dynamics on the SMPL body model.
Our experiments reveal that pose noise is amplified by ${\sim}1{,}000\times$ into torque error, that proximal joints are up to $10\times$ more sensitive than distal joints, and that low-pass filtering and differentiable pose refinement can substantially reduce this amplification.
These findings suggest that for downstream physics applications, trunk estimation accuracy matters far more than limb accuracy, and that future pose estimation benchmarks should incorporate physics-aware metrics alongside kinematic accuracy.
The differentiable nature of our module opens a natural path forward, as it can be directly integrated into existing pose estimator training pipelines as a physics-informed loss, enabling end-to-end optimization for both kinematic accuracy and dynamic plausibility.

{
    \small
    \bibliographystyle{ieeenat_fullname}
    \bibliography{main}
}
\end{document}